\documentclass[lettersize,journal]{IEEEtran}
\usepackage{amsmath,amsfonts}
\usepackage{algorithmic}
\usepackage{algorithm}
\usepackage{array}
\usepackage[caption=false,font=normalsize,labelfont=sf,textfont=sf]{subfig}
\usepackage{textcomp}
\usepackage{stfloats}
\usepackage{url}
\usepackage{verbatim}
\usepackage{graphicx}
\usepackage{cite}
\usepackage{multirow}
\usepackage{xcolor}

\newcommand\boldp[1]{\vspace{.1cm}\noindent\textbf{#1.}\hspace{.065cm}}
\begin{document}

\title{SGCNeRF: Few-Shot Neural Rendering via Sparse Geometric Consistency Guidance}

\author{
Yuru Xiao, Xianming Liu,~\IEEEmembership{Member,~IEEE,} Deming Zhai,~\IEEEmembership{Member,~IEEE,} Kui Jiang,~\IEEEmembership{Member,~IEEE,} Junjun Jiang,~\IEEEmembership{Senior Member,~IEEE,} Xiangyang Ji,~\IEEEmembership{Member,~IEEE}
\IEEEcompsocitemizethanks{
\IEEEcompsocthanksitem  Y. Xiao, X. Liu, D. Zhai, K. Jiang and J. Jiang are with the School of Computer Science and Technology, Harbin Institute of Technology, Harbin, China, E-mail: xiaoyuru.30@gmail.com; \{csxm, zhaideming, jiangkui, jiangjunjun\}@hit.edu.cn.
\IEEEcompsocthanksitem X. Ji is with the Department of Automation, Tsinghua University,
Beijing, 100084, China, E-mail: xyji@tsinghua.edu.cn.
% \IEEEcompsocthanksitem This work was supported by National Natural Science Foundation of China (xxxx). 
} 

}
% The paper headers
\markboth{IEEE Transactions on Circuits and Systems for Video Technology}%
% {Shell \MakeLowercase{\textit{et al.}}: A Sample Article Using IEEEtran.cls for IEEE Journals}
{Xiao \MakeLowercase{\textit{et al.}}: SGCNeRF: Few-Shot Neural Rendering via Sparse Geometric Consistency Guidance}

\maketitle

\begin{abstract}

Neural Radiance Field (NeRF) technology has made significant strides in creating novel viewpoints. However, its effectiveness is hampered when working with sparsely available views, often leading to performance dips due to overfitting. FreeNeRF attempts to overcome this limitation by integrating implicit geometry regularization, which incrementally improves both geometry and textures. Nonetheless, an initial low positional encoding bandwidth results in the exclusion of high-frequency elements. The quest for a holistic approach that simultaneously addresses overfitting and the preservation of high-frequency details remains ongoing. This study presents a novel feature-matching-based sparse geometry regularization module, enhanced by a spatially consistent geometry filtering mechanism {and a frequency-guided geometric regularization strategy}. This module excels at accurately identifying high-frequency keypoints, effectively preserving fine structural details. Through progressive refinement of geometry and textures across NeRF iterations, we unveil an effective few-shot neural rendering architecture, designated as SGCNeRF, for enhanced novel view synthesis. Our experiments demonstrate that SGCNeRF not only achieves superior geometry-consistent outcomes but also surpasses FreeNeRF, with improvements of 0.7 dB in PSNR on LLFF and DTU.
\end{abstract}

\begin{IEEEkeywords}
Few-shot neural rendering, Feature match,  Frequency regularization.
\end{IEEEkeywords}

\section{Introduction}
\label{sec:intro}

\IEEEPARstart{N}{eural} Radiance Field (NeRF) \cite{mildenhall2021nerf} has garnered significant attention in the fields of computer vision and computer graphics owing to its exceptional performance in %the task of 
novel view synthesis (NVS). While NeRF has exhibited commendable realism in neural rendering, its superior performance in reconstruction is heavily contingent upon the availability of dense input views. In real-world scenarios, such as those encountered in robotics \cite{blukis2023oneshot}, autonomous driving \cite{ziyang2023snerf,chen2021geosim}, and outdoor environments \cite{wang2023digging,rematas2022urban}, the input view is often sparse, making them insufficient for reconstructing a 3D scene due to the propensity for overfitting. Consequently, the quality of the generated novel view is far from meeting the practical application requirements. 

\begin{figure}[!t]
    \centering
    \includegraphics[width=1\linewidth]{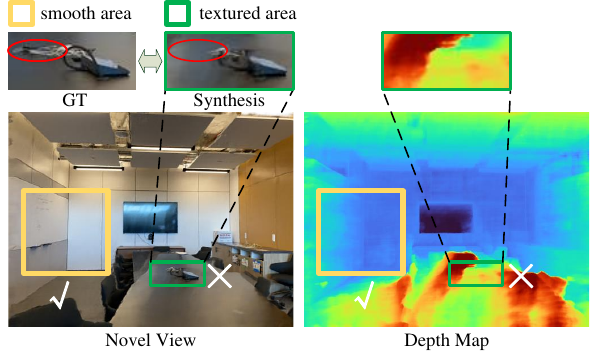}

   \caption{ \textbf{%Example FreeNeRF Results
   Visual results of FreeNeRF.}
    NeRF with frequency regularization have exhibited remarkable performance in regions with low texture or geometry variation (yellow box). However, %their efficacy significantly decreases 
    they struggle to generate visual pleasing results in areas with rich textures or geometry variation (green box).}
   \label{fig:motivation}
\end{figure}

The exploration of few-shot neural rendering has uncovered a multitude of strategies aimed at confronting the ubiquitous challenge of overfitting. Among these, RegNeRF \cite{niemeyer2022regnerf} stands out with its patch-based geometry regularization technique. Despite its efficacy, it  encounters under-constrained artifacts due to geometric inconsistency. Alternatively, learning-based methods {\cite{lin2023vision,yu2021pixelnerf,mvsnerf, chibane2021stereo,zhang2025nerfprior, ni2024colnerf}} engage expansive datasets to train neural networks, often resulting in blurred outputs due to relatively loose geometric constraints unless further fine-tuned. Diffusion-based approaches \cite{chen2023single,gu2023nerfdiff,shue20233d,xu2022sinnerf} have also been introduced, showcasing their ability to produce detailed 3D radiance fields from singular views, while their effectiveness presently remains confined to simple, object-centric scenes.

A pivotal area of ongoing research within few-shot neural rendering is the integration of geometric consistency. DS-NeRF \cite{deng2022depth} attempts to achieve this by adding a sparse 3D point cloud, obtained via structure-from-motion algorithms, though its effectiveness is limited by the sparsity of point extraction from extremely limited input views. While some techniques \cite{wang2023sparsenerf,roessle2022dense} employ depth estimation or completion networks to secure dense depth maps for improved supervision, challenges persist in upholding geometric consistency, largely owing to the intrinsic ambiguities in processing multiple inputs with these networks. The advent of deep matcher-based methods \cite{truong2023sparf,lao2024corresnerf} signifies progress by extracting geometry priors from dense correspondences, yet they still struggle with blurring and the precise rendering of fine details due to inaccuracies in correspondences.

Given the aforementioned context, the quest for a few-shot neural rendering methodology that not only ensures high geometric consistency, especially in high-frequency regions, but also adeptly combats overfitting continues to be a vigorous and ongoing endeavor.
It is remarkable that, FreeNeRF \cite{yang2023freenerf} showcases a promising approach by progressively increasing the frequency of positional encoding, a technique referred to as frequency regularization. This method has proven particularly effective in areas dominated by low frequencies, as depicted by the regions highlighted in yellow in Fig. \ref{fig:motivation}, leveraging its intrinsic smoothing properties. Nonetheless, this approach encounters a notable decline in performance concerning both the visual quality of the rendered images and the maintenance of geometric consistency within high-frequency regions, as delineated by the segments highlighted in green in Fig. \ref{fig:motivation}. Such observations motivate the exploration of a specialized methodology aimed at enhancing the geometric consistency of high-frequency components, thus addressing the nuanced demands of few-shot neural rendering.

Drawing inspiration from the principle of sparse image matching, in this paper, we introduce an innovative sparse geometric consistency method, termed as SGCNeRF, for few-shot novel view synthesis. 
The key parts of SGCNeRF involve sparse feature match, frequency regularization and explicit geometry regularization. 
Initially, we utilize a pre-trained sparse matching operator to establish correspondences on high-frequency keypoints offline. These corresponding pixels are then mapped into 3D space based on their rendered depth. By minimizing the distances between the paired 3D points, we comprehensively optimize the geometry of the radiance field. This process simultaneously enforces a robust constraint on the placement of high-frequency keypoints to prevent excessive smoothing caused by the frequency regularization method.
Acknowledging the potential decline in performance caused by inaccurate correspondences, we propose a simple yet effective filter that ensures geometric consistency, which excludes inconsistent correspondences, defined as rays with a minimum distance exceeding a predefined threshold, significantly improving overall geometric consistency. {To further alleviate distortions resulting from matching errors in the post-training stage, we introduce a frequency-guided geometry regularization scheme, which adaptively relaxes geometric constraints through a frequency annealing schedule, promoting more stable and accurate geometry refinement.}
Overall, our proposed SGCNeRF method achieves sparse geometric consistency in high-frequency areas while maintaining the superior performance of frequency regularization  in regions with low texture or geometry variation. Extensive experiments  demonstrate that our method outperforms the state-of-the-art methods on multiple datasets, without necessitating additional 3D data or training interventions. 

The main contributions of our work are summarized as follows:

\begin{itemize}
    \item We introduce a novel perspective that accurately locating sparse keypoints while gradually refining is a more effective strategy for few-shot neural rendering, especially in recovering fine details.

    \item We reveal and explore the potential complementarity between the sparse matcher and frequency regularization method, which fully harmonizes the merits of high geometric consistency and the inherent smoothing characteristics within the context of few-shot neural rendering.

    \item {We propose a frequency-guided geometric regularization method tailored to our sparse localization and progressive refinement framework. This approach effectively leverages geometric priors derived from the matching network, while simultaneously mitigating distortions introduced by matching errors.}

\end{itemize}

\section{Related Work}
\label{relate}
\boldp{Neural Radiance Fields} 
Neural Radiance Fields (NeRF) \cite{mildenhall2021nerf} model scenes as continuous density and color functions. The density fields provide information about radiance and transmittance, while the color fields indicate illuminated textures. In recent years, NeRF has gained considerable attention and has been applied in various areas of research \cite{mildenhall2022nerf,wang2022nerf}, such as novel view synthesis \cite{barron2021mip,barron2022mip}, scene relighting \cite{rudnev2022nerf,zhi2022dual}, {style transfer \cite{wang2025stylizing}, radiance field compression \cite{yang2025shell}}, and dynamic field \cite{liu2023robust,fang2022fast}. However, the commendable performance of NeRF is notably contingent upon the availability of dense input views. Consequently, achieving a NeRF representation of superior quality through sparse inputs remains a formidable challenge.

\boldp{Few-shot Neural Rendering} 
Numerous scholarly endeavors have been undertaken to tackle the challenge posed by the few-shot neural rendering problem. One prevalent approach involves incorporating additional 3D supervision, such as sparse point clouds \cite{deng2022depth, rematas2022urban, wang2023neural} or dense depth maps \cite{roessle2022dense, wang2023sparsenerf}, to regularize the geometry. In addition to extra 3D supervision, RegNeRF \cite{niemeyer2022regnerf} employs a smooth prior on random patches to optimize geometry, while InfoNeRF \cite{kim2022infonerf} utilizes a simple ray entropy regularization method to mitigate overfitting.
More recently, the FreeNeRF \cite{yang2023freenerf} method has employed an implicit geometry regularization method named frequency regularization to address the overfitting issue in the context of few-shot neural rendering. Similarly, training methodologies characterized by a coarse-to-fine progression have found application in diverse domains, including bundle adjustment \cite{lin2021barf}, surface reconstruction \cite{wang2022hf, huang2023scneus}, and deformable fields \cite{park2021nerfies}.
Existing methods primarily address overfitting concerns but tend to overlook performance in high-frequency detail areas. This study endeavors to identify a comprehensive solution that effectively addresses both issues. 

\boldp{Image Matching in Few-shot Neural Rendering} 
Recently, SPARF \cite{truong2023sparf} has effectively addressed the challenges associated with sparse and noisy camera poses, benefiting from an image matcher \cite{truong2021learning}. However, the efficacy of SPARF is contingent upon the quantity and precision of correspondences. In contrast, our framework is designed to prioritize enhancing geometric consistency within high-frequency components, particularly within the constraints of sparse view settings with fixed ground truth poses. This orientation signifies that our approach does not depend on an abundance of correspondences in low-frequency areas, but places greater emphasis on accuracy within the high-frequency areas.

\section{Preliminaries and Motivation}
\label{Preliminaries}

NeRF \cite{mildenhall2021nerf} employs a multi-layer perceptron (MLP) to represent a density field $\sigma$ and a color field $\mathbf{c}$ as a continuous implicit function $f_\theta(\mathbf{x}, \mathbf{d}) = (\sigma, \mathbf{c})$. Here, $\mathbf{x} \in \mathbb{R}^3$ denotes an input 3D coordinate, and $\mathbf{d} \in \mathbb{R}^3$ indicates the ray direction. A point on a ray $\mathbf{r} = (\mathbf{o}, \mathbf{d})$ can be expressed as $\mathbf{p} = \mathbf{o} + t\mathbf{d}$, where $t$ represents the distance from the origin $\mathbf{o}$. NeRF randomly samples $N$ points $\{t_1, ..., t_N\}$ along the ray within the specified range of $[t_n, t_f]$. The color of the ray can be rendered using the volume rendering equation, depicted as
\begin{equation}
\label{eq:vr}
\hat{\mathbf{c}}(\mathbf{r}) = \sum_{i=1}^{N}T(t_i)(1-\exp{(-\sigma(t_i)\delta(t_i))})\mathbf{c}(t_i),
\end{equation}
where $T(t_i)=\exp{(-\sum_{j=1}^{i-1}\sigma(t_j)\delta(t_j))}$ represents the transmittance from the nearest point to the sampled point $\mathbf{o}+t_i\mathbf{d}$, and $\delta(t_i)=t_{i+1} - t_i$ denotes the distance between samples. Moreover, the approximate depth $\hat{z}$ along the ray can be obtained as
\begin{equation}
    \label{eq:depth}
    \hat{z}(\mathbf{r}) = \sum_{i=1}^{N}T(t_i)(1-\exp{(-\sigma(t_i)\delta(t_i))})t_i.
\end{equation}
To better represent high-frequency details, NeRF introduces  positional encoding technique to map 3D positional vectors into a higher-dimensional space. The positional encoding function is expressed as 
\begin{equation}
\mathbf{\gamma}(\mathbf{x}) = (\sin(2^0 \mathbf{x}),\cos(2^0 \mathbf{x}),...,\sin(2^{L-1} \mathbf{x}),\cos(2^{L-1} \mathbf{x})),
\end{equation}
where $L$ represents the hyperparameter that determines the maximum frequency of the encoding.

\boldp{Frequency Regularization}  
FreeNeRF \cite{yang2023freenerf} provides an alternative solution to the few-shot neural rendering problem by employing a concise line of code to gradually increase the frequency of the encoding. Specifically, the formulaic descriptions of this technique %involves the following step:
can be depicted as
\begin{equation}
    \gamma^\prime(t, T, \mathbf{x}) = \gamma(\mathbf{x}) \odot \mathbf{M}(t, T, L)
    \label{eq:ec}
\end{equation}
with
\begin{equation}
    \mathbf{M}_i(t, T, L) = \begin{cases}1 & \text{if } i \le \frac{t \cdot L}{T} + 3 \\ \frac{t \cdot L}{T} - \lfloor\frac{t \cdot L}{T}\rfloor & \text{if } \frac{t \cdot L}{T} + 3 < i \le \frac{t \cdot L}{T} + 6 \\ 0 & \text{if } i > \frac{t \cdot L}{T} + 6\end{cases},
    \label{eq:mask}
\end{equation}
where $t$ and $T$ respectively refer to the current iteration step and the total iteration step; $\gamma$ and $\gamma^\prime$ separately denotes the initial positional encoding and masked positional encoding; $i$ represents the index along the $L$-dimensional axis. 
%$\gamma$ is the initial positional encoding, and 
$\mathbf{M}$ is the frequency mask. The frequency of positional encoding is gradually increased in a linear fashion during the training process.
\begin{figure}[!htp]
   \centering
  % \fbox{\rule{0pt}{2in} \rule{0.9\linewidth}{0pt}}
   \includegraphics[width=1\linewidth]{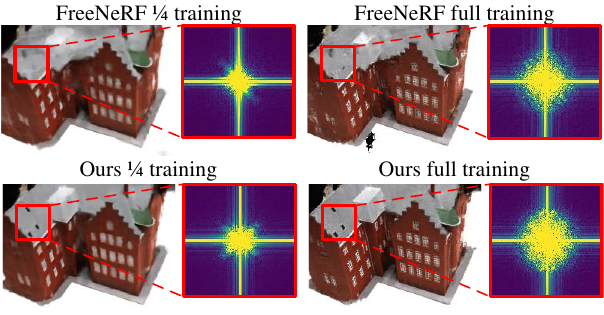}

   \caption{\textbf{Mitigating high-frequency truncation during initial training.} FreeNeRF exhibits an over-smoothing in its early training stages, resulting in the loss of fine details. In contrast, our proposed method introduces constraints to mitigate this over-smoothing effect, preserving the intricate details. }
   \label{fig:truncation}
\end{figure}

\begin{figure*}[!htp]
   \centering
  % \fbox{\rule{0pt}{2in} \rule{0.9\linewidth}{0pt}}
   \includegraphics[width=1\linewidth]{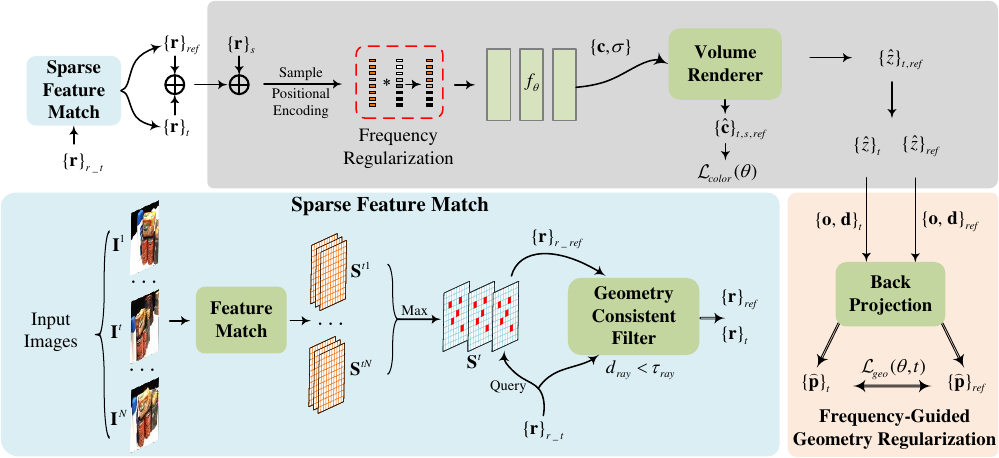}

   \caption{\textbf{Overview of our proposed sparse geometric consistency neural rendering framework (SGCNeRF).} Our proposed method aims to guide the recovery of fine details, which %may be smoothed during a coarse-to-fine training process
   are commonly lost in existing technologies \cite{yang2023freenerf}. Specifically, we employ a sparse matcher \cite{lindenberger2023lightglue} to establish correspondences on high-frequency areas across multiple views. Subsequently, we use a geometry-consistent filter to eliminate imperfect correspondences. The remaining paired rays are then mapped to 3D space using the rendered depth. To ensure the gradual recovery of high-frequency details, we enforce geometry consistency by minimizing the distance between paired 3D points.}  %The rendered depths of these paired rays are used to calculate the geometry regularization loss. }
   \label{fig:method}
\end{figure*}
\boldp{Motivation}  
During the initial stages of the training process, it is crucial to acknowledge that the rendered novel view may exhibit over-smoothing, as shown in the small red box in the left portion of Fig. \ref{fig:truncation}. This phenomenon can be attributed to the low frequency of positional encoding, a result of employing frequency regularization (refer to Eq. \ref{eq:ec}). The significant reduction in frequency bandwidth, exemplified by the notable diminution in the large red box, further emphasizes the issue of high-frequency truncation.
As training progresses, the frequency of positional encoding gradually increases. However, it is noteworthy that the lost details from the initial stages are not fully recovered, as depicted in the right portion of  Fig. \ref{fig:truncation}. This observation serves as motivation for mitigating high-frequency truncation, particularly in detailed areas, during the initial training phase. Once the over-smoothed structures are coarsely recovered in the early stages, further refinement of these structures can be achieved in subsequent training iterations (see Fig. \ref{fig:truncation}). 

\section{Method}
\label{Method}
In the concluding section, we delve into an in-depth analysis to unravel the underlying reasons for FreeNeRF's \cite{yang2023freenerf} shortcomings. The investigation discerns {frequency truncation during initial training} as the primary factor responsible. In response to this,  we introduce a novel sparse geometry regularization approach meticulously crafted to {mitigate frequency truncation by identifying high-frequency keypoints}.  Utilizing a sparse feature-matching network, we establish correspondences across multiple input views and enforce geometry regularization on the resulting pairs, aligning with the principles of multi-view geometry. The details of the feature-matching method are expounded in \ref{matching}. To ensure the accuracy of correspondences, we introduce a straightforward geometry-consistent filter that eliminates inaccurate correspondences through the analytical computation of the minimal distance between two rays. This filter significantly improves performance and is detailed in \ref{Filter}. Finally, {we propose a frequency-guided geometric regularization loss as outlined in \ref{Loss},} and the visualization of our method is presented in Fig. \ref{fig:method}.

\subsection{Sparse Feature Matching}
\label{matching}
Our method for sparse geometry regularization relies on establishing robust correspondences within regions of high frequency. To accomplish this, we employ the {LightGlue} network \cite{lindenberger2023lightglue}, a pre-trained model recently introduced and celebrated for its proficiency in aligning keypoints identified by the {Super-Point} algorithm \cite{detone2018superpoint} with high effectiveness.

In scenarios that involve processing multiple input views, we categorize each image as either the target image $\mathbf{I}^t$ or as one of the reference images $\{\mathbf{I}^r\},$ where $r$ belongs to the set $\{1, \ldots, N\}$ and $r \neq t$. Subsequently, we form image pairs such as $(\mathbf{I}^t,\mathbf{I}^1),\ldots,(\mathbf{I}^t,\mathbf{I}^N)$ and feed these pairs into the LightGlue network. This process generates a series of matching stacks $\mathbf{S}^{t1},\ldots,\mathbf{S}^{tN}$ for all the paired images.
Each matching stack is composed of three channels, which respectively represent the corresponding pixels' image index, pixel index, and the confidence in their match (with dimensions specified as HWC). To amalgamate all these matching stacks into a unified stack, we implement a strategy that prioritizes matches based on the principle of maximum confidence:
\begin{equation}
    \label{aggregate}
    \mathbf{S}^{t}_{i,j,k}=\mathbf{S}^{tz^\ast}_{i,j,k},\:\:z^\ast = \max_z \mathbf{S}^{tz}_{i,j,3},
\end{equation}
{where $i$, $j$, and $k$ represent the indices of the tensor along the height, width, and channel (HWC) dimensions.} respectively. The principle dictates that each keypoint within an image is linked to precisely one corresponding ray with maximum confidence. This strategy aims to prevent the occurrence of multiple rays aligning with a singular keypoint. Such a scenario could compromise the effectiveness of our regularization methodology, particularly when these multiple rays are mapped to different three-dimensional coordinates during joint NeRF optimization. To ensure the efficient training of the NeRF backbone, we perform the feature matching process offline to acquire the matching stack $\mathbf{S}^t, t \in \{1,...,N\}$ for all input images. In each training iteration, we randomly sample rays $\{\mathbf{r}\}_{r\_t}$ from a randomly selected image $\mathbf{I}^t$. We then retrieve the corresponding rays $\{\mathbf{r}\}_{r\_ref}$ from the stack $\mathbf{S}^t$.

\subsection{Geometry Consistent Filter}
\label{Filter}
Sparse matchers, such as those described in \cite{lindenberger2023lightglue, sarlin2020superglue, Chen_2021_ICCV}, have demonstrated remarkable effectiveness in synchronously matching features and filtering out outliers. However, these tools exhibit vulnerabilities in scenarios characterized by repeated objects or uniform surfaces. These challenges can undermine the dependability of the matching process, which in turn, may adversely affect the performance of our geometry regularization approach.

To address these concerns, we devise a simple yet effective method to assess the consistency of correspondences within the 3D space. This method is founded on the premise that a true correspondence should originate from a singular point within the scene, suggesting that the corresponding rays should converge at that exact location. This is quantified by ensuring that the minimum distance between the two rays is negligible, essentially approaching zero. Although this criterion alone is not entirely conclusive, it provides a basis for eliminating ray pairs that exhibit a significant minimum distance between them. Specifically, given the origins $(\mathbf{o}_1, \mathbf{o}_2)$ and directions $(\mathbf{d}_1, \mathbf{d}_2)$ of the paired rays, we calculate the minimum distance $d_{ray}$ between these rays by
\begin{equation}
    d_{ray} = \|\mathbf{o}_1-\mathbf{o}_2+m\mathbf{d}_1-n\mathbf{d}_2\|_2,
    % \begin{aligned} 
    %     d_{ray} &= \|\mathbf{P}_1-\mathbf{P}_2\|_2 \\
    %     \mathbf{P}_1-\mathbf{P}_2 &= \mathbf{o}_1-\mathbf{o}_2+m\mathbf{d}_1-n\mathbf{d}_2
    % \end{aligned},
    \label{eq:dray}
\end{equation}
where $m$ and $n$ represent the distances from the origin points to the corresponding points with minimum distance in the two rays. 
To establish the formulation of $m$ and $n$, we utilize a fundamental theorem in elementary geometry: ``The shortest distance segment connecting two non-coplanar lines in three-dimensional space is the perpendicular segment that links the two lines." This theorem can be expressed as
\begin{equation}
\label{eq:theorem}
    \left\{
    \begin{aligned}
    \mathbf{d}_1\cdot(\mathbf{P}_1-\mathbf{P}_2) &= 0 \\
    \mathbf{d}_2\cdot(\mathbf{P}_1-\mathbf{P}_2) &= 0
    \end{aligned},
    \right.
\end{equation}
where  $\mathbf{d}_1$ and $\mathbf{d}_2$ denote the direction vectors of the two rays, while $\mathbf{P}_1=\mathbf{o}_1+m\mathbf{d}_1$ and $\mathbf{P}_2=\mathbf{o}_2+n\mathbf{d}_2$ represent the two endpoints of the line segment. By substituting the values of $\mathbf{P}_1$ and $\mathbf{P}_2$ into Eq. \ref{eq:theorem}, we simplify the equation to get Eq. \ref{eq:simlified}:
\begin{equation}
    \label{eq:simlified}
    \left\{
    \begin{aligned}
    m\mathbf{d}_1\cdot\mathbf{d}_2 - n\mathbf{d}_1\cdot\mathbf{d}_2 &= \mathbf{d}_1\cdot\mathbf{o}_2 - \mathbf{d}_1\cdot\mathbf{o}_1 \\
    m\mathbf{d}_1\cdot\mathbf{d}_2-n\mathbf{d}_2\cdot\mathbf{d}_2 &= \mathbf{d}_2\cdot\mathbf{o}_2-\mathbf{d}_2\cdot\mathbf{o}_1
    \end{aligned}.
    \right.   
\end{equation}
By solving Eq. \ref{eq:simlified}, we derive the expressions for the variables $m$ and $n$, which are represented as

\begin{equation}
    \label{eq:dist}
    \begin{aligned}
        m &= \frac{be+cf-cg-bd}{ab-cc} \\
        n &= \frac{ce+af-cd-ag}{ab-cc},
    \end{aligned}
\end{equation}
where
\begin{equation}
    \begin{aligned}
        a &= \mathbf{d}_1 \cdot \mathbf{d}_1 \quad
        b = \mathbf{d}_2 \cdot \mathbf{d}_2 \quad
        c = \mathbf{d}_1 \cdot \mathbf{d}_2 \\
        d &= \mathbf{d}_1 \cdot \mathbf{o}_1 \quad
        e = \mathbf{d}_1 \cdot \mathbf{o}_2 \quad
        f = \mathbf{d}_2 \cdot \mathbf{o}_1 \\
        g &= \mathbf{d}_2 \cdot \mathbf{o}_2.
    \end{aligned}
\end{equation}

We discard paired rays whose distance exceeds the threshold, $d_{ray}>\tau_{ray}$, and forward the remaining rays, $\{\mathbf{r}\}_{t,ref}$, to the subsequent NeRF backbone for the computation of geometry regularization loss.

\subsection{Frequency-Guided Geometry Regularization}
\label{Loss}
{Our central claim is that high-precision localization during the initial training stage facilitates the construction of a radiance field that captures the global structure. This coarse representation can later be refined through pixel-wise photometric loss. However, directly introducing sparse geometric constraints in the subsequent training stage—when high-frequency inputs are incorporated—often leads to distortions caused by feature matching errors, even with outlier filtering. These errors mainly stem from misalignments between corresponding rays. To address this issue and fully leverage the benefits of pixel-wise photometric supervision, we propose a frequency-guided geometry regularization loss. This loss progressively relaxes geometric constraints over the course of training, following a frequency annealing schedule defined as:}
\begin{equation}
\label{eq:geoloss}
\mathcal{L}_{geo}(\theta,t) = \lambda_{geo}\omega_f(t)\sum_{i=1}^{N}||\hat{\mathbf{p}}_t^i-\hat{\mathbf{p}}_{ref}^{i}||_2,
\end{equation}
{where $\hat{\mathbf{p}}_t^i \in \hat{\mathbf{p}}_t$ and $\hat{\mathbf{p}}_{\text{ref}}^i \in \hat{\mathbf{p}}_{\text{ref}}$ denote the $i$-th pair of corresponding 3D points. Here, $N$ denotes the number of correspondences, $t$ represents the current iteration step, and $\lambda_{\text{geo}}$ is a hyperparameter fixed at 0.1 throughout all experiments. The \textit{frequency-guided weighting function} $\omega_f(t)$ is computed based on the current frequency mask $\mathbf{M}(t, T, L)$ as:}
\begin{equation}
    \omega_f(t) = 2^{\lambda_f(1-\frac{\Sigma_j \mathbf{M(t,T,L)}}{\Sigma_j \mathbf{M(0,T,L)}})},
\end{equation}
{where $\lambda_f$ is a hyperparameter that controls the decay rate of the frequency weighting function, and $j$ denotes the index of elements in the frequency mask. By incorporating the proposed frequency-guided geometry regularization, we effectively leverage the geometric prior provided by the sparse matching network during the initial phase of training, while mitigating distortions caused by feature matching errors in the subsequent stages.}

\section{Experiments}
\boldp{Datasets and Metrics}
We evaluate our method on three real-world multi-view datasets, namely LLFF \cite{mildenhall2019local}, DTU \cite{jensen2014large}{, and MipNeRF360 \cite{barron2022mip}}. The LLFF dataset~\cite{mildenhall2019local} consists of eight forward-facing scenes. Following FreeNeRF~\cite{yang2023freenerf}, we assign every eighth image as the novel view for evaluation. To construct input views, we uniformly sample from the remaining images. For image downsampling, we apply an average filter with a reduction factor of 8. {The MipNeRF360 dataset consists of seven large-scale scenes. For evaluation, we adopt the test split protocol used in the LLFF dataset. A downsampling factor of 4 is applied following CoR-GS \cite{zhang2024cor}.}

The DTU dataset~\cite{jensen2014large} is a comprehensive multi-view dataset comprising 124 object-centric scenes. Following the PixelNeRF approach~\cite{yu2021pixelnerf}, we partition the dataset into 88 training scenes and 15 test scenes, identified by the scan IDs 8, 21, 30, 31, 34, 38, 40, 41, 45, 55, 63, 82, 103, 110, and 114.
Beyond the standard partitioning, we conduct a qualitative assessment to validate our results across alternative scenarios within the DTU dataset. We specifically focus on scan IDs 24, 37, and 106. For training within each scan scene, we utilize image IDs 25, 22, 28, 40, 44, 48, 0, 8, and 13. The initial three IDs are chosen for the three-input view setting, respectively.
The remaining images, with IDs 1, 2, 9, 10, 11, 12, 14, 15, 23, 24, 26, 27, 29, 30, 31, 32, 33, 34, 35, 41, 42, 43, 45, 46, and 47, are reserved for evaluation purposes. Consistent with RegNeRF~\cite{niemeyer2022regnerf} and FreeNeRF~\cite{yang2023freenerf}, we employ an average filter to downsample images with a reduction factor of 4.

For quantitative results, we report the PSNR, SSIM, and LPIPS scores for view synthesis. All quantitative results are averaged among all test views in each dataset following \cite{yang2023freenerf}.

\begin{figure*}[t]
   \centering

   \includegraphics[width=1\linewidth]{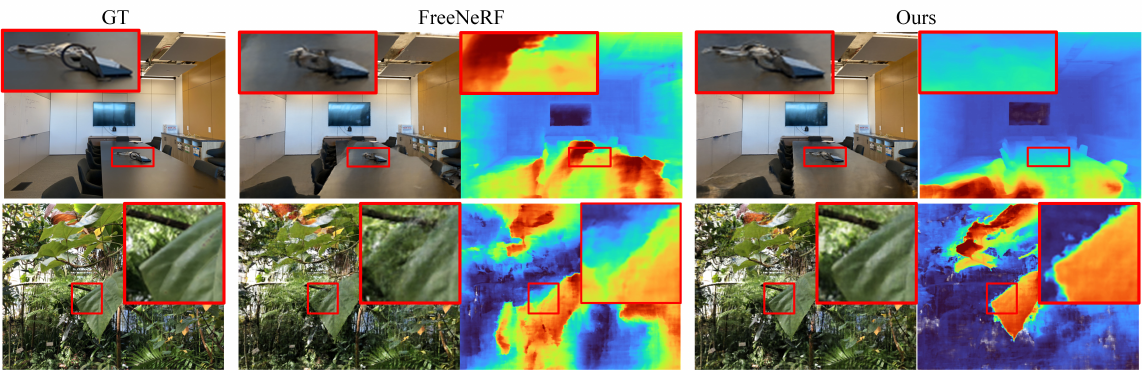}

   \caption{\textbf{Qualitative comparison on LLFF.} We present novel views and depth maps rendered by FreeNeRF and our approach in a 3-view setting. FreeNeRF demonstrates limitations in certain areas, as highlighted by the red boxes, where the rendered outputs display absent or distorted structures. In contrast, our method surpasses FreeNeRF in these regions by adeptly leveraging sparse geometric consistency constraints.}
   \label{fig:LLFF}
\end{figure*}

\boldp{Baselines}
Unless given specific instructions, we adopt the results presented in FreeNeRF \cite{yang2023freenerf}, since our implementation is based on its codebase. Specifically, we compare our method with works focused on few-shot neural rendering, including PixelNeRF \cite{yu2021pixelnerf}, MVSNeRF \cite{mvsnerf}, RegNeRF \cite{niemeyer2022regnerf}, DietNeRF \cite{jain2021putting}, DSNeRF \cite{deng2022depth}, GeCoNeRF \cite{kwak2023geconerf}, SparseNeRF\cite{wang2023sparsenerf} and FreeNeRF \cite{yang2023freenerf}, along with the standard Mip-NeRF \cite{barron2021mip}. Additionally, we conduct a comparison with the most recent few-shot NeRF method, SPARF \cite{truong2023sparf}, which employs a dense matcher and a coarse-to-fine training strategy.
Notably, SPARF incorporates Gaussian Pyramid Downsampling during the preprocessing of the DTU dataset and utilizes AlexNet for computing LPIPS metrics. In contrast, our approach employs interpolation for downsampling and leverages VGG for LPIPS metric computation. As a result, we replicate the LPIPS metrics across all datasets, while reproducing all quantitative results on the preprocessed DTU dataset with fixed ground truth poses consistent with our methodology.

\boldp{Settings}
Our approach follows the FreeNeRF settings, with the exception of our geometry regularization module. For optimization, we employ the Adam optimizer with a learning rate that exponentially decays from $2\cdot10^{-3}$ to $2\cdot10^{-5}$. During the initial 512-step training, a warm-up strategy with a multiplier of 0.01 is implemented. To ensure stability, gradients are clipped by value at 0.1 and normalized to 0.1. Specifically, we train the network for 500 epochs with a batch size of 4096. This results in 44k training iterations on the DTU dataset~\cite{jensen2014large} for 3 input views and 70k training iterations for those on the LLFF dataset~\cite{mildenhall2019local}. In our experiments, the threshold parameter $\tau_{ray}$ for the geometry regularization module is set to 0.15 for the LLFF dataset and 0.0017 for the DTU dataset. 

\boldp{Sampling Strategy}
We observe a significant disparity between the quantity of correspondences generated by the feature-matching module and the number of pixels present in an image. To address this issue, we utilize the original photometric supervision in regions without correspondences. This involves applying a standard FreeNeRF~\cite{yang2023freenerf} procedure to a randomized selection of rays $\{\mathbf{r}\}_s$ from these regions.
{In each iteration step, we consistently set the number of target rays ($n_t$) or reference rays ($n_{ref}$) to 50. This ensures that the ratio of sampled ray pairs to the total number of correspondences in an image approximately matches the ratio of the pixel batch size to the total number of pixels, thereby maintaining balanced training. A substantial increase in the number of correspondences used at each step disrupts the overall balance and results in a slight degradation in performance, as shown in Tab.~\ref{tab:nc}.} Additionally, to maintain consistency with other baselines, we fix the total number of rays dispatched to the renderer backbone at 4096, equivalent to the batch size of other baselines. Consequently, there is a constant number of rays $\{\mathbf{r}\}_s$, specifically 3996, for all our experiments.

\begin{table}[!htp]
\caption{Effect of Varying the Number of Correspondences at Each Step.}\label{tab:nc}
\centering
\scalebox{1}{
\begin{tabular}{c|ccc}  \hline
 $n_{t}$ &  PSNR$\uparrow$ & SSIM$\uparrow$ & LPIPS$\downarrow$ \\ \hline

 200    &  {20.15}          &  {0.650}          &  0.288  \\

 50    &  {20.33} &  {0.661} & {0.276}       \\ \hline

\end{tabular}}
\end{table}

\boldp{Occlusion Regularization}
FreeNeRF~\cite{yang2023freenerf} incorporates a straightforward occlusion regularization term to address potential ``floater'' or ``walls'' artifacts that may arise when the number of training views is limited. This technique constrains the density in the near camera area to zero, proving effective even in regions where geometry regularization cannot be applied due to the absence of correspondences. Consequently, we have adopted the occlusion regularization term as our default choice.
The final training objective is formulated as $\mathcal{L}(\theta,t) = \mathcal{L}_{color}(\theta) + \mathcal{L}_{geo}(\theta,t) + \lambda_{occ}\mathcal{L}_{occ}(\theta)$, where $\lambda_{occ}$ is pre-defined weighting factors. Throughout all experiments, we set the value of $\lambda_{occ}$ to 0.01. For a more comprehensive understanding of $\mathcal{L}_{occ}$, we recommend referring to FreeNeRF~\cite{yang2023freenerf}.

\subsection{Comparison on the LLFF Dataset}
We conduct a comprehensive analysis of both qualitative and quantitative results using the LLFF dataset \cite{mildenhall2019local}, as depicted in Fig. \ref{fig:LLFF} and Tab. \ref{tab:llff}, respectively. In the qualitative assessment, FreeNeRF \cite{yang2023freenerf} demonstrates commendable performance in low-frequency areas like walls and ceilings but exhibits failures in specific detailed areas, as highlighted by the red boxes. In contrast, our method, leveraging sparse geometry consistency guidance, achieves superior results in high-frequency areas.
Quantitatively, our approach significantly outperforms FreeNeRF across various metrics, especially in scenarios with sparser input views (e.g., surpassing FreeNeRF by 0.7 dB in PSNR with 3 input views).
Additionally, our method surpasses the most recent few-shot NeRF technique, which includes a dense matcher, achieving higher performance metrics (e.g., exceeding SPARF by 0.13 dB in PSNR and by 0.03 in SSIM). The notable increase in SSIM demonstrates our method's superior capability in reconstructing fine details.

\begin{table}[!t]
\caption{{Quantitative results on LLFF with 3 input views.} \label{tab:llff}}
\centering

\begin{tabular}{l||l|ccc}  \hline
\multirow{2}{*}{Method} & \multirow{2}{*}{Venue} & \multicolumn{3}{c}{LLFF} \\ &  & PSNR$\uparrow$ & SSIM$\uparrow$ & LPIPS$\downarrow$ \\ \hline
PixelNeRF \cite{yu2021pixelnerf} & CVPR·21 & 7.93 & 0.272 & 0.682  \\
MVSNeRF \cite{mvsnerf} & ICCV·21  & 17.25          & 0.557          & 0.356     \\
mip-NeRF \cite{barron2021mip} & ICCV·21 & 14.62          & 0.351          & 0.495      \\
DietNeRF \cite{jain2021putting} & ICCV·21  & 14.94          & 0.370          & 0.496        \\ 
DSNeRF \cite{deng2022depth} & CVPR·22  & 18.94          & 0.582          & 0.362        \\ 
RegNeRF \cite{niemeyer2022regnerf} & CVPR·22  & \textbf{19.08}         & \textbf{0.587}          & \textbf{0.336}       \\ \hline

FreeNeRF \cite{yang2023freenerf} & CVPR·23 &  {19.63}          &  {0.612}          & 0.308  \\

SparseNeRF \cite{wang2023sparsenerf} & ICCV·23    &  {19.86}          &  {0.624}          &  0.328  \\

GeCoNeRF \cite{kwak2023geconerf} & ICML·23    &  {18.77}          &  {0.596}          &  0.338  \\

{CorresNeRF \cite{lao2023corresnerf}} &  {NIPS·23}    &   {19.83}          &   {0.700}          &   {0.290}  \\

3DGS \cite{kerbl20233d} & TOG·23    &  {14.94}          &  {0.468}          &  0.379  \\

SPARF \cite{truong2023sparf} & CVPR·23    &  {20.20}          &  {0.630}          &  0.327  \\
DNGaussian \cite{li2024dngaussian} & CVPR·24    &  {19.12}          &  {0.591}          &  {0.294}  \\

Ours  & --    &  \textbf{20.33} &  \textbf{0.661} & \textbf{0.276}       \\ \hline

\end{tabular}
\vspace{-2em}
\end{table}

\subsection{Comparison on the DTU Dataset}
\begin{figure*}[t]
   \centering

   \includegraphics[width=1\linewidth]{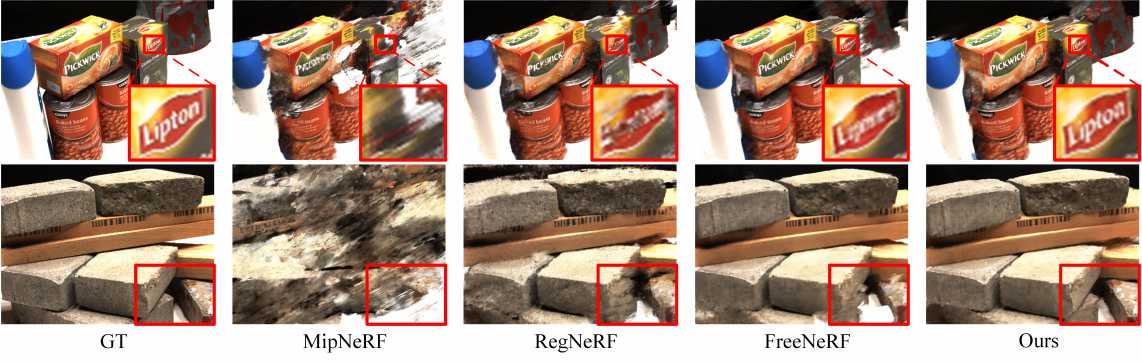}

   \caption{\textbf{Qualitative comparison on DTU.} We show novel views rendered by MipNeRF, RegNeRF, FreeNeRF, as well as our method utilizing three input views. The baseline methods exhibit limitations in rendering detailed areas, such as letters and bricks.}
   \label{fig:DTU}
\end{figure*}

To facilitate a more comprehensive evaluation, we conduct both qualitative and quantitative comparisons using the DTU dataset \cite{jensen2014large}, as illustrated in Fig. \ref{fig:DTU} and Tab. \ref{tab:dtu}, respectively. Our qualitative results involves comparisons with MipNeRF \cite{barron2021mip}, RegNeRF \cite{niemeyer2022regnerf}, and FreeNeRF \cite{yang2023freenerf}. For instance, the baseline methods fail to render the high-frequency details (highlighted by red boxes), such as letters in the first row, and bricks in the second row, while our method effectively addresses such issues by constraining the geometry from being over-smoothed by the high-frequency truncation. 
In terms of quantitative results, our method outperforms the latest baseline FreeNeRF \cite{yang2023freenerf} and SPARF \cite{truong2023sparf}, especially when dealing with sparser input views, such as three input views (surpassing FreeNeRF and SPARF by 0.7 dB and 1.0 dB in terms of PSNR, respectively). It is worth noting that although PixelNeRF \cite{yu2021pixelnerf} yields better PSNR outcome than other baselines in the DTU dataset with three input views, this is due to the learned geometry prior on the white and black background trained on the DTU dataset \cite{yang2023freenerf}. In contrast, our method does not require additional training processes, making it more efficient and robust.

\begin{table}[!t]
\caption{{Quantitative results on DTU with 3 input views.\label{tab:abl}} \label{tab:dtu}}
\centering
\scalebox{0.92}{
\begin{tabular}{l||l|ccc}  \hline
\multirow{2}{*}{Method} & \multirow{2}{*}{Setting} & \multicolumn{3}{c}{LLFF} \\ &  & PSNR$\uparrow$ & SSIM$\uparrow$ & LPIPS$\downarrow$ \\ \hline
PixelNeRF \cite{yu2021pixelnerf} & Pretrain + Finetune & \textbf{18.74} & \textbf{0.618} & 0.401  \\
MVSNeRF \cite{mvsnerf} & Pretrain  + Finetune & 16.33          & 0.602          & \textbf{0.385}     \\ \hline
mip-NeRF \cite{barron2021mip} & Anti-aliasing & 7.64          & 0.227          & 0.655      \\
DietNeRF \cite{jain2021putting} & Semantic Consistency  & 10.01          & 0.354          & 0.574        \\ 
RegNeRF \cite{niemeyer2022regnerf} & Geometry Regularization  & 15.33          & 0.621          & 0.341       \\ 

FreeNeRF \cite{yang2023freenerf} & Frequency Regularization &  {18.02}          &  {0.680}          & {0.318}  \\

SPARF \cite{truong2023sparf} & Dense Matcher    &  {17.75}          &  {0.681}          &  0.328  \\

Ours  & Sparse Matcher + Freq.Reg    &  \textbf{18.71} &  \textbf{0.691} & \textbf{0.310}       \\ \hline
\end{tabular}}
\end{table}

\subsection{Comparison on the MipNeRF360 Dataset}
\begin{figure*}
    \centering
    \includegraphics[width=1\linewidth]{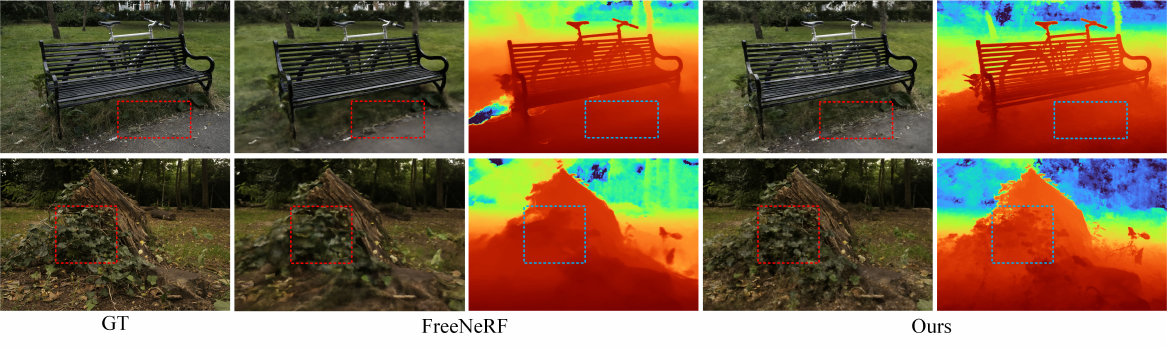}
    \caption{\textbf{Qualitative Comparison on MipNeRF360.} We compare our method with FreeNeRF on the MipNeRF360 dataset using 24 input views. Both methods are implemented based on the MipNeRF360 architecture to ensure a fair comparison.}
    \label{fig:360}
\end{figure*}

{To rigorously assess the effectiveness of our method in large-scale, real-world scenarios, we perform experiments on the MipNeRF360 dataset \cite{barron2022mip} using the MipNeRF360\cite{barron2022mip} framework. Fig. \ref{fig:360} and Tab. \ref{tab:360} report the qualitative and quantitative results, respectively. Quantitative comparisons indicate that our method outperforms both FreeNeRF \cite{yang2023freenerf} and the more recent CoR-GS \cite{zhang2024cor}, exceeding the latter by 0.4 dB in terms of PSNR. Qualitatively, our approach yields more detailed and visually accurate renderings compared to FreeNeRF. Notably, FreeNeRF tends to produce overly smoothed RGB and depth images—an issue effectively addressed by our method.
}

\begin{figure*}[t]
   \centering

   \includegraphics[width=0.9\linewidth]{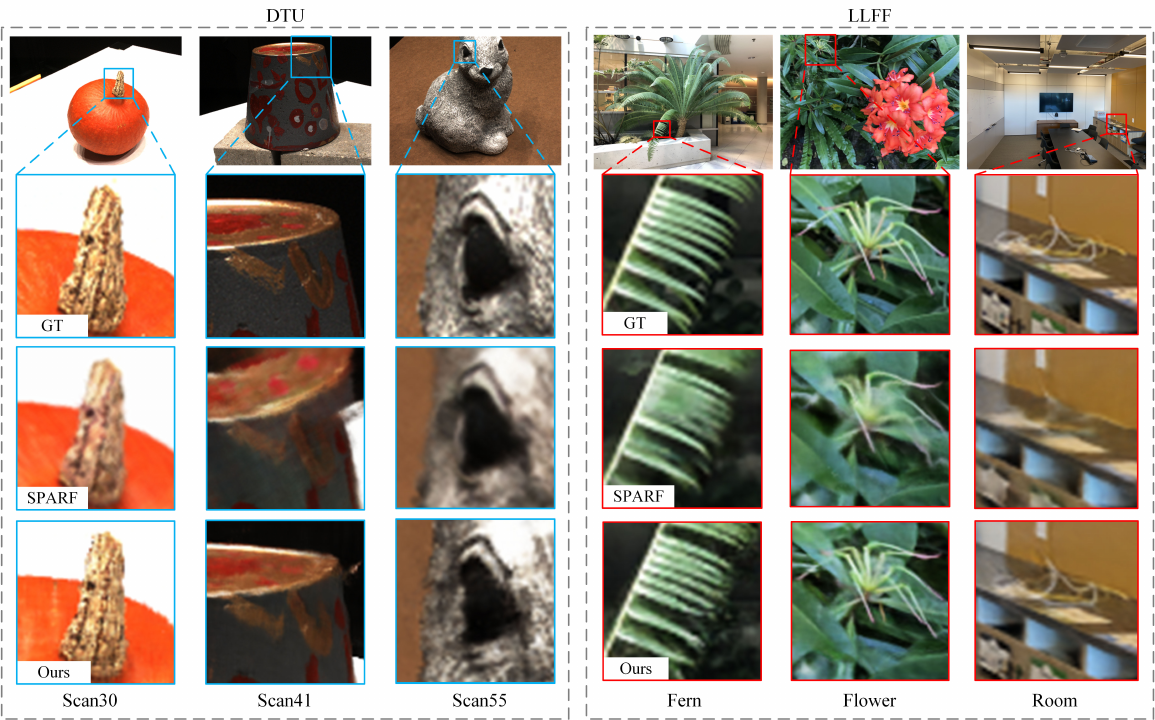}

   \caption{\textbf{Qualitative comparison with SPARF on DTU and LLFF.} We offer a qualitative comparison of our method with the latest dense matcher-based technique, SPARF \cite{truong2023sparf}. Our findings indicate superior performance in high-frequency areas.}
   \label{fig:LLFf_DTU}
\end{figure*}

\begin{table}[!t]
\caption{{Quantitative results on MipNeRF360 with 24 input views.} \label{tab:360}}
\centering

\begin{tabular}{l||l|ccc}  \hline
\multirow{2}{*}{Method} & \multirow{2}{*}{Venue} & \multicolumn{3}{c}{MipNeRF360} \\ &  & PSNR$\uparrow$ & SSIM$\uparrow$ & LPIPS$\downarrow$ \\ \hline
FreeNeRF \cite{yang2023freenerf} & CVPR·21 & 22.69 & 0.598 & 0.424  \\
3DGS \cite{kerbl20233d} & TOG·23 & 22.80 & 0.708 & 0.276  \\
CoR-GS \cite{zhang2024cor} & ECCV·24 & 23.39 & \textbf{0.727} &  \textbf{0.271}  \\
Ours  & --    &  \textbf{23.84} &  {0.696} & {0.293}       \\ \hline

\end{tabular}
% \vspace{-2em}
\end{table}
\boldp{Comparison with Dense Matcher-based Methods}
Previous techniques have integrated dense matchers into few-shot neural rendering, as documented in \cite{truong2023sparf,lao2024corresnerf}, capitalizing on the pre-trained priors of dense matching networks. This approach effectively addresses under-constrained geometry challenges. However, the accuracy limitations of dense matchers have resulted in blurriness in high-frequency areas, as illustrated in Fig. \ref{fig:LLFf_DTU}. In contrast, our approach pioneers the use of a sparse matching network in few-shot neural rendering, achieving more precise correspondences. By incorporating frequency regularization, our method maximizes the use of photometric loss to gradually refine high-frequency details, provided that high-frequency areas are precisely located. The qualitative and quantitative results presented in Fig. \ref{fig:LLFf_DTU} and Tab. \ref{tab:densecom} demonstrate our method's enhanced performance in both low-frequency and fine-detail areas, respectively.

\begin{table}[!t]
\caption{A comparison between our method and dense matcher-based approaches.} 
\label{tab:densecom}
\centering

\begin{tabular}{l||l|cccc}  \hline
\multirow{2}{*}{Method} & \multirow{2}{*}{Venue} & \multicolumn{4}{c}{LLFF} \\
 &  & PSNR$\uparrow$ & SSIM$\uparrow$ & LPIPS$\downarrow$  & Train$\downarrow$ \\ \hline
CorresNeRF \cite{lao2024corresnerf}  & NIPS·23    & 19.83  & \textbf{0.700}& 0.290 &  \textbf{3h}    \\
SPARF \cite{truong2023sparf}  & CVPR·23    & 20.20  & 0.630& 0.327 &  13h      \\

Ours  & --    &  \textbf{20.33} &  0.661&\textbf{0.276} & \textbf{3h}      \\ \hline

\end{tabular}
% \vspace{-2em}
\end{table}

\subsection{Effectiveness for the 3DGS framework}
{Recently, 3D Gaussian {S}platting (3DGS) {\cite{kerbl20233d, liu2025review}} has achieved remarkable performance and enabled efficient training and rendering, thanks to its explicit Gaussian representation and differentiable rasterization. In this pipeline, Gaussians generated from a sparse point cloud are progressively refined using the position gradient derived from the photometric loss. In this paper, we introduce a novel approach that combines sparse localization with gradual refinement and incorporates a geometric consistency filter—a strategy that is also applicable to 3DGS-based methods. }

\begin{figure}[!htp]

   \centering

   \includegraphics[width=1\linewidth]{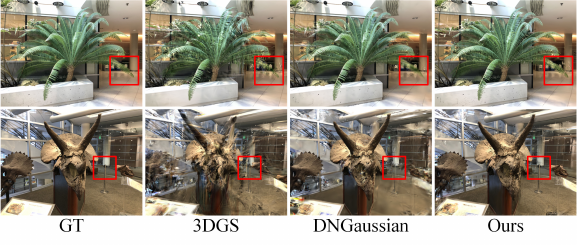}

   \caption{We evaluated our method using the 3DGS architecture, and our results demonstrate that it captures finer details compared to both the base 3DGS architecture and DNGaussian.}
   \label{fig:3dgs}
\end{figure}

{We conducted experiments to validate the effectiveness of our approach using the 3DGS architecture. {We implement a sparse matching-based geometry regularization method—without frequency guidance—built upon the original 3DGS framework, and adopt the sparse-view setting as proposed in DNGaussian \cite{li2024dngaussian}.} The qualitative and quantitative results are presented in Fig. \ref{fig:3dgs} and Tab. \ref{tab:3dgs}, respectively. In these experiments, we integrate mid-points derived from the correspondences with the randomly initialized points and apply geometric regularization to the correspondences filtered by our outlier filter. We compare our method with both the base 3DGS method and the sparse-view 3DGS method, DNGaussian \cite{li2024dngaussian}, with both baselines initialized using random points. Our approach substantially enhances the performance of the 3DGS base architecture in handling sparse-view scenarios. By incorporating the sparse matcher and our geometry-consistent filter, the proposed method achieves a notable improvement of 3 dB over the baseline 3DGS architecture.}

\begin{table}[!t]
\caption{Quantitative results on LLFF with 3 input views based on 3DGS architecture.}
\label{tab:3dgs}
\centering
\scalebox{1}{
\begin{tabular}{l||l|ccc}  \hline
\multirow{2}{*}{Method} & \multirow{2}{*}{Setting} & \multicolumn{3}{c}{LLFF} \\ &  & PSNR$\uparrow$ & SSIM$\uparrow$ & LPIPS$\downarrow$ \\ \hline
3DGS\cite{kerbl20233d}  & - & {16.50} & {0.476} & 0.354  \\
DNGaussian\cite{li2024dngaussian}  & - & {19.12} & {0.591} & 0.294  \\ \hline
\multirow{5}{*}{Ours w/3DGS}  & $\tau = 0.05$    &  {19.46} &  {0.659} & {0.226}       \\ 
  & $\tau = 0.10$    &  {19.75} &  {0.671} & {0.218}       \\ 
  & $\tau = 0.20$    &  {19.91} &  \textbf{0.677} & \textbf{0.213}       \\ 
  & $\tau = 0.40$    &  \textbf{19.93} &  \textbf{0.677} & {0.214}       \\ 

& $\tau = \infty$    &  19.90 &  {0.646} & {0.237}       \\ \hline
\end{tabular}}
\end{table}

\subsection{Method Analysis}

\begin{figure*}[!htp]
   \centering
  % \fbox{\rule{0pt}{2in} \rule{0.9\linewidth}{0pt}}
   \includegraphics[width=1\linewidth]{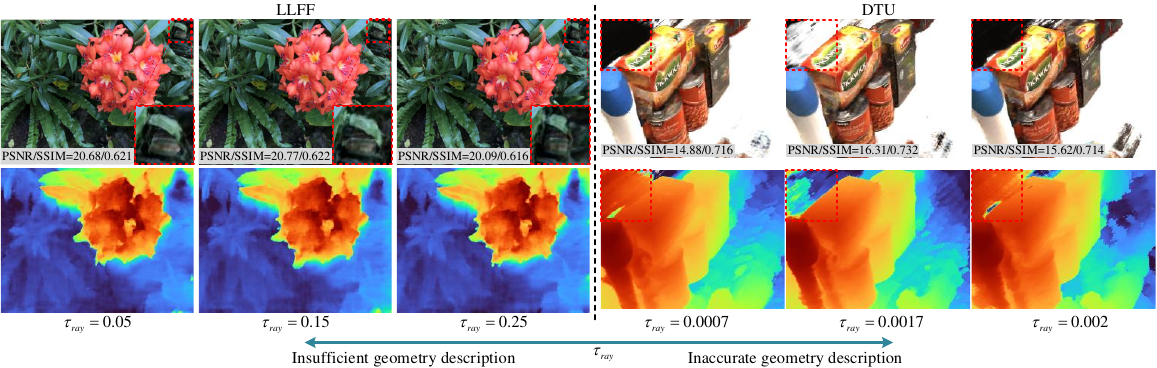}

   \caption{We present qualitative ablations exploring the impact of the geometry-consistent filter on the LLFF dataset (left portion) and the DTU dataset (right portion). For the LLFF dataset, we vary $\tau_{ray}$ from 0.05 to 0.25, while for the DTU dataset, we vary $\tau_{ray}$ from 0.0007 to 0.002. The qualitative results indicate that extremely small or large $\tau_{ray}$ values cause distorted geometry. This distortion arises from either insufficient or inaccurate geometry descriptions, respectively. The ablation study further verifies the importance of the geometry-consistent filter when applying feature-matching-based geometry regularization.}
   \label{fig:abl_dist}
\end{figure*}

\boldp{Impact of Geometry-consistent Filter}
Tab. \ref{tab:filter_thr} illustrates the ablation study conducted to investigate the impact of the threshold $\tau_{ray}$ on matching accuracy. For the LLFF dataset, $\tau_{ray}$ varies from 0.05 to 0.35 in increments of 0.05. It is observed that while the SSIM and LPIPS metrics exhibit minimal changes, the PSNR metric fluctuates around its optimal value. In contrast, for the DTU dataset, characterized by smaller scene scales compared to LLFF, the optimal PSNR value ranges from 0.0005 to 0.002. Here, $\tau_{ray}$ varies from 0.0005 to 0.002, with step sizes of approximately 0.0002 or 0.0003. Similar to the LLFF dataset, slight variations are observed in the SSIM and LPIPS metrics, while the PSNR metric exhibits significant fluctuations. These observations suggest that $\tau_{ray}$ strongly influences the PSNR metric, which evaluates overall consistency. Higher values lead to more inaccurate correspondences, while lower values result in less detailed geometry descriptions (refer to Fig. \ref{fig:abl_dist}), both negatively affecting the PSNR metric.
\begin{table}[!t]
    \caption{Impact of threshold $\tau_{ray}$}
    \label{tab:filter_thr}
    \centering
    \scalebox{0.9}{
    \begin{tabular}{l|ccc||l|ccc} \hline
        \multirow{2}{*}{$\tau_{ray}$} & \multicolumn{3}{c||}{LLFF} & \multirow{2}{*}{$\tau_{ray}$} &  \multicolumn{3}{c}{DTU} \\
             & PSNR$\uparrow$ & SSIM$\uparrow$ & LPIPS$\downarrow$  &  & PSNR$\uparrow$ & SSIM$\uparrow$ & LPIPS$\downarrow$ \\ \hline
        0.05 & 19.77 & 0.639 & 0.289  & 0.0005  & 18.18 & 0.679 & 0.323 \\
        0.1 & 20.21 & 0.658 & 0.278  & 0.0007 & 18.01 & 0.684 & 0.324 \\
        0.15 & \textbf{20.33} & \textbf{0.661} &  \textbf{0.276} & 0.001 & 18.33 & 0.688 & 0.318 \\ 
        0.2 &  20.27 &  \textbf{0.661} & 0.277 & 0.0012 & 17.96 & 0.679 & 0.326 \\
        0.25 & 20.21 & \textbf{0.661} & 0.277  & 0.0015 & 18.22 & 0.690 & 0.318 \\
        0.3 & 20.25 & {0.660} & {0.278}  & 0.0017 & \textbf{18.71} & \textbf{0.691} & \textbf{0.310} \\ 
        0.35 & 20.28 & {0.660} & 0.277  & 0.002 & 18.34 & 0.683 & 0.321 \\ \hline

    \end{tabular}}

\end{table}

Consequently, we conclude that the effectiveness of our sparse feature matching-based geometry regularization heavily relies on the geometric consistency between correspondences. Optimizing performance is feasible even in scenarios with a scarcity of correspondences, provided that such pairs exhibit a high degree of geometric consistency. This observation underscores the critical significance of geometric consistency in the context of sparse matching-based few-shot neural rendering.

\boldp{Ablation of Various Matching Network}
{Image matching is a fundamental computer vision task for 3D reconstruction and camera calibration. In recent years, robust end-to-end networks have been developed, including sparse matchers (such as SuperGlue\cite{sarlin2020superglue} and LightGlue\cite{lindenberger2023lightglue}) and dense matchers (such as RoMa \cite{edstedt2024roma}, DUSt3R \cite{wang2024dust3r}, and MASt3R \cite{leroy2024grounding}). When comparing these approaches, sparse matchers typically offer faster matching speeds and higher accuracy, while dense matchers capture fine-grained details in low-texture areas, making them more suitable for complete dense reconstruction.}

{Our work addresses the loss of high-frequency details resulting from frequency regularization. To overcome this issue, we choose LightGlue—a sparse matcher—to accurately locate high-frequency features while avoiding the disturbances in low-texture regions that may arise from the less precise matching of dense methods. We argue that once low-texture areas are well reconstructed through frequency regularization, additional geometric regularization from dense matchers can have a detrimental effect. In contrast, sparse matchers, which establish correspondences only at high-frequency keypoints, are more effective in correcting the original issues caused by frequency regularization.}

{To validate our claim, we conducted experiments using various matching networks. As shown in Tab. \ref{tab:matching}, even the state-of-the-art dense matcher RoMa is incompatible with the frequency regularization strategy and performs significantly worse compared to the integration of sparse matchers such as SuperGlue or LightGlue. This discrepancy arises because the coarse representation during initial training is highly sensitive to the accuracy of correspondences, and a large number of low-accuracy matches can adversely affect the initial geometry reconstruction.}

\begin{table}[!t]
\caption{Quantitative ablation of various matching network.}
\label{tab:matching}
\centering
\scalebox{0.85}{
\begin{tabular}{l||l|cccc}  \hline
\multirow{2}{*}{Matcher} & \multirow{2}{*}{Setting} & \multicolumn{4}{c}{LLFF} \\ &  & PSNR$\uparrow$ & SSIM$\uparrow$ & LPIPS$\downarrow$ & $\mathbb{E}(d_{ray})$ \\ \hline
\multirow{4}{*}{Ours w/RoMa \cite{edstedt2024roma}}  & $\tau = 0.05$ & 19.77 & 0.641 &  0.293 & 0.0298 \\
  & $\tau = 0.15$    &  20.01  & 0.657  & 0.279  & 0.0381      \\ 
    & $\tau = 0.25$    & 19.99  &  0.655 & 0.285   & 0.0382     \\ 
  & $\tau = 0.35$    & 19.82  & 0.637  & 0.289    & 0.0383   \\ \hline

\multirow{4}{*}{Ours w/SuperGlue \cite{sarlin2020superglue}}  & $\tau = 0.05$ & 20.10 & 0.648 & 0.289  & 0.0231 \\
  & $\tau = 0.15$    & 20.13  & 0.652  & 0.277  & 0.0338     \\ 
    & $\tau = 0.25$    &  20.17 & 0.655  & 0.279   & 0.0343     \\ 
  & $\tau = 0.35$    & 20.21  &  0.662 & 0.270    & 0.0345    \\ \hline

  \multirow{4}{*}{Ours w/LightGlue \cite{lindenberger2023lightglue}}  & $\tau = 0.05$ & 19.77 & 0.639 & 0.289 & 0.0250 \\
  & $\tau = 0.15$    & \textbf{20.33}  & \textbf{0.661}  & \textbf{0.276}  & 0.0335     \\ 
    & $\tau = 0.25$    &  20.21 & \textbf{0.661} & 0.277 & 0.0335       \\ 
  & $\tau = 0.35$    & 20.28 & {0.660} & 0.277    & 0.0336    \\ \hline

\end{tabular}}
\end{table}

\begin{figure}
    \centering
    \includegraphics[width=1\linewidth]{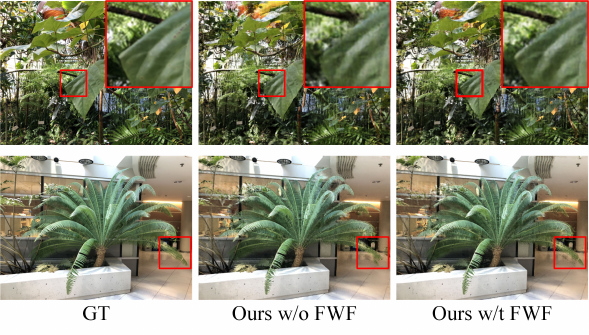}
    \caption{We present a qualitative analysis of the proposed Frequency-Guided Weighting Function (FWF).}
    \label{fig:frequency_weight}
\end{figure}

\boldp{Ablation of Frequency-guided Weighting Function}
{In this paper, we propose a frequency-guided geometry regularization strategy designed to better leverage geometric priors from the matching network while mitigating the impact of matching errors during the subsequent training process. Quantitative and qualitative analyses are presented in Tab. \ref{tab:abl} and Fig. \ref{fig:frequency_weight}, respectively. The quantitative results demonstrate that applying the geometric loss function without frequency guidance leads to a noticeable performance drop—approximately 0.15 dB in PSNR. The qualitative results further confirm that incorporating the frequency-guided weighting function (FWF) yields rendered outputs with finer details and reduced distortion.}

\begin{table}[!htp]
\caption{{Ablation Study.} \label{tab:abl}}
\centering
\scalebox{0.85}{
\begin{tabular}{l||ccc|ccc} \hline
\multirow{2}{*}{Setting} & \multicolumn{3}{c|}{LLFF} & \multicolumn{3}{c}{DTU} \\ 
 &  PSNR$\uparrow$ & SSIM$\uparrow$ & LPIPS$\downarrow$ & PSNR$\uparrow$ & SSIM$\uparrow$ & LPIPS$\downarrow$ \\ \hline
w/o. Occlusion Reg. & 20.23 & 0.651 & 0.288 & 16.31 & 0.638 & 0.379 \\
w/o. Frequency Reg. & 19.68 & 0.647 & 0.290 & 14.23 & 0.557 & 0.399\\
w/o. Geometry Reg.  & 19.63 & 0.612 & 0.308 & 18.02 & 0.680 & {0.318} \\
w/o. Filter         & 20.15 & 0.659 & 0.290 & 18.22 & 0.688 & 0.320\\
% Ours (w/t. NeRF)          & 20.14 & 0.657 & \textbf{0.283} & 18.19 & 0.669  & 0.347 \\
w/o. FWF        & 20.18 & 0.657 & 0.280 & 18.42 & 0.690 & 0.314\\
Ours             & \textbf{20.33} & \textbf{0.661} & \textbf{0.276} & \textbf{18.71} & \textbf{0.691} & \textbf{0.310} \\ \hline

\end{tabular}}

\end{table}

\boldp{Ablation Study}
We conduct a comprehensive ablation study on the LLFF and DTU datasets, using three input views, as shown in Tab.~\ref{tab:abl}, {based on the MipNeRF architecture}. 
A noteworthy observation is the significant performance decline of our method in the absence of frequency regularization. Conversely, the full method demonstrates superior outcomes. This discrepancy arises from the intentionally sparse design of correspondences, aimed at ensuring geometric consistency in high-frequency areas but lacking the ability to address overfitting in other areas. Therefore, the sparse geometry regularization functions more as a guide, ensuring that the coarse-to-fine training with the assurance that high-frequency keypoints are accurately located.

\section{Conclusion}
In this paper, we introduced a novel sparse geometry regularization module that leverages the advantages of an innovative geometry-consistent filter {and a frequency-guided geometry regularization strategy}, thereby ensuring high geometric consistency. In-depth scrutiny is conducted to underscore the significance of sparse geometric consistency in addressing the limitations of conventional frequency regularization methods, particularly in handling failure cases pertaining to high-frequency components. The proposed module establishes a distinctive framework designed to tackle the challenges inherent in few-shot neural rendering. Specifically, it serves as a sparse geometry regularization method for precise localization of high-frequency keypoints and as a frequency regularization method for the smooth refinement of low-frequency structures. Empirical findings attest to the complementary contributions of these two methods, collectively yielding a state-of-the-art performance in few-shot neural rendering.

\bibliographystyle{IEEEtran}

\bibliography{main}

\section{Biography Section}
% If you have an EPS/PDF photo (graphicx package needed), extra braces are
%  needed around the contents of the optional argument to biography to prevent
%  the LaTeX parser from getting confused when it sees the complicated
%  $\backslash${\tt{includegraphics}} command within an optional argument. (You can create
%  your own custom macro containing the $\backslash${\tt{includegraphics}} command to make things
%  simpler here.)
 
% \vspace{11pt}

% \bf{If you include a photo:}
\vspace{-33pt}
\begin{IEEEbiography}[{\includegraphics[width=1in,height=1.25in,clip,keepaspectratio]{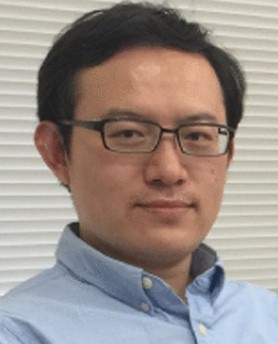}}]{Xianming Liu}
(Member, IEEE) received the BS, MS, and PhD degrees in computer science from HIT, in 2006, 2008 and 2012, respectively. He is a professor with the School of Computer Science and Technology, Harbin Institute of Technology (HIT), Harbin, China. 
In 2011, he spent half a year with the Department of Electrical and Computer Engineering, McMaster University, Canada, as a visiting student, where he then worked as a post-doctoral fellow from 2012 to 2013. He worked as a project researcher with the National Institute of Informatics (NII), Tokyo, Japan, from 2014 to 2017. He has published more than 60 international conference and journal publications, including top IEEE journals, such as IEEE Transactions on Image Processing, IEEE Transactions on Circuits and Systems for Video Technology, IEEE Transactions on Information Forensics and Security, IEEE Transactions on Multimedia, IEEE Transactions on Geoscience and Remote Sensing; and top conferences, such as CVPR, IJCAI and DCC. He is the receipt of IEEE ICME 2016 Best Student Paper Award.
\end{IEEEbiography}
\vspace{-33pt}
\begin{IEEEbiography}[{\includegraphics[width=1in,height=1.25in,clip,keepaspectratio]{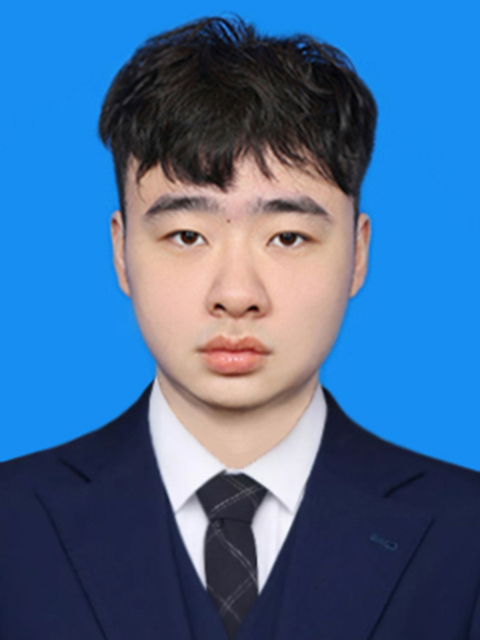}}]{Yuru Xiao}
received M.Eng. degree in 2022 from Harbin Institute of Technology, China. He is now pursuing a Ph.D. degree at the School of Computer Science and Technology, Harbin Institute of Technology. His current research focuses on 3D vision, computer graphics and neural rendering.
\end{IEEEbiography}

\vspace{-33pt}

\begin{IEEEbiography}[{\includegraphics[width=1in,height=1.25in,clip,keepaspectratio]{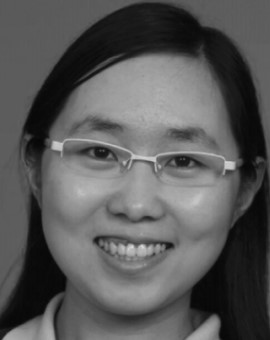}}]{Deming Zhai}
(Member, IEEE) received the B.S., M.S., and Ph.D. (Hons.) degrees in computer science from the Harbin Institute of Technology (HIT), Harbin, China, in 2007, 2009, and 2014, respectively. She is currently an Associate Professor with the Department of Computer Science, HIT. 
In 2011, she was with The Hong Kong University of Science and Technology, as a Visiting Student. In 2012, she was with the GRASP Laboratory, University of Pennsylvania, PA, USA, as a Visiting Scholar. From August 2014 to April 2016, she was a Project Researcher with the National Institute of Informatics (NII), Tokyo, Japan.
\end{IEEEbiography}
\vspace{-33pt}
\begin{IEEEbiography}[{\includegraphics[width=1in,height=1.25in,clip,keepaspectratio]{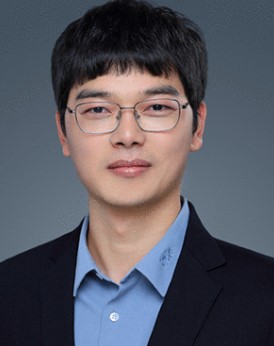}}]{Kui Jiang}
(Member, IEEE) received the Ph.D. degree from the School of Computer Science, Wuhan University, Wuhan, China, in 2022. He is currently an Associate Professor with the School of Computer Science and Technology, Harbin Institute of Technology, Harbin, China. 
His research interests include image/video processing and computer vision. He was the recipient of the 2023 CSIG Excellent Doctoral Dissertation Award and 2022 ACM Wuhan Doctoral Dissertation Award.
\end{IEEEbiography}
\vspace{-33pt}
\begin{IEEEbiography}[{\includegraphics[width=1in,height=1.25in,clip,keepaspectratio]{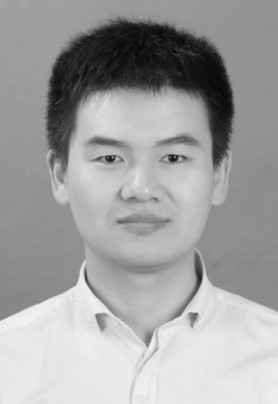}}]{Junjun Jiang}
(Senior Member, IEEE) received the B.S. degree in Information and Computing Science from Huaqiao University, Quanzhou, China, in 2009, and the Ph.D. degree in Computer Science and Technology from Wuhan University, Wuhan, China, in 2014. He is currently a Professor with the School of Computer Science and Technology, Harbin Institute of Technology, Harbin, China. 
He received the 2016 China Computer Federation (CCF) Outstanding Doctoral Dissertation Award and 2015 ACM Wuhan Doctoral Dissertation Award, His research interests include image processing and computer vision.
\end{IEEEbiography}
\vspace{-33pt}
\begin{IEEEbiography}[{\includegraphics[width=1in,height=1.25in,clip,keepaspectratio]{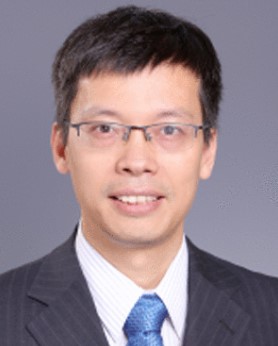}}]{Xiangyang Ji}
(Member, IEEE) received the BS degree in materials science and the MS degree in computer science from the Harbin Institute of Technology, Harbin, China, in 1999 and 2001, respectively, and the PhD degree in computer science from the Institute of Computing Technology, Chinese Academy of Sciences, Beijing, China. He joined Tsinghua University, Beijing, in 2008, where he is currently a professor with the Department of Automation, School of Information Science and Technology. 
He has authored more than 100 referred conference and journal papers. His current research interests include signal processing, image/video compressing, and intelligent imaging.
\end{IEEEbiography}

\vspace{-33pt}

\vfill

\end{document}